\documentclass[a4paper,12pt]{article}
\usepackage[american]{babel}
\usepackage{apacite}

\newif\ifpdf
\ifx\pdfoutput\undefined
   \pdffalse 
\else
   \pdfoutput=1
   \pdftrue
\fi

\ifpdf
   \usepackage[pdftex]{graphicx}
   \usepackage[OT1]{fontenc}
   \usepackage[isolatin]{inputenc}

\else
   \usepackage{graphicx}
   \usepackage[OT1]{fontenc}
   \usepackage[isolatin]{inputenc}
   
\fi
\usepackage{amsmath,amssymb}
\usepackage{array}
\usepackage{supertabular}

\usepackage{amsmath}
\usepackage{amssymb}
\usepackage{setspace}
\title{Spontaneous Dynamics of Asymmetric Random Recurrent Spiking Neural Networks}
\author{H\'edi Soula$^1$, Guillaume Beslon$^1$, Olivier Mazet$^2$ \\
$^1$Artificial Life and Behavior -- PRISMA \\
$^2$Applied Mathematics Lab -- MAPLY \\ 
National Institute of Applied Science \\ 
LYON -- FRANCE }
\date{October 6, 2004}
\newcommand{\E}{\mathbb{E}}

\begin{document}


\maketitle
\thispagestyle{empty}
\abstract{
We study in this paper the effect of an unique initial stimulation on random recurrent networks of leaky integrate and fire neurons. Indeed given a stochastic connectivity this so-called spontaneous mode exhibits various non trivial dynamics. This study brings forward a mathematical formalism that allows us to examine the variability of the afterward dynamics according to the parameters of the weight distribution. Provided independence hypothesis (e.g. in the case of very large networks) we are able to compute the average number of neurons that fire at a given time -- the spiking activity. In accordance with numerical simulations, we prove that this spiking activity reaches a steady-state, we characterize this steady-state and explore the transients.



}
\pagestyle{empty}
\section{Introduction}%
Many modern neurobiological problems have to confront the behaviors of large recurrent spiking neuron networks. Indeed, it is assumed that these observable behaviors are a result of a collective dynamics of interacting neurons. So the question that rises becomes : given a connectivity of the network and a single neuron property, what are the possible kinds of dynamics~?
\par
In the case of  homogeneous nets (same connectivity inside the network), some authors found sufficient conditions for phase synchronization (locking) or stability \cite{chow98,gerstner01b}. Another author \cite{coombes99} calculated Lyapunov exponents in a given symmetric connectivity map and showed that some neurons were ``chaotic'' (i.e. the highest exponent was positive). 
In the very general case \cite{golomb94,vanvreeswijk96,meyer02} it has been shown that the dynamics can show a broad variety of aspects. 
\par
In the particular case of integrate and fire neurons, \cite{amit97,amit97a} used consistency techniques on nets of irregular firing neurons. This technique allowed them to derive a self-sustaining criterion. Using Fokker-Planck diffusion, the same kind of methods was used in the case of linear I\&F neurons in \cite{fusi99,mattia00}, for stochastic networks dynamics with noisy input current \cite{guidice03} and in the case of sparse weight connectivity \cite{brunel99}.
\par
However, stochastic recurrent spiking neurons networks are rarely studied in their spontaneous functioning. Indeed most the time the dynamics is driven by an external current -- whether meaningful or noisy. Nevertheless, without this external current, the resulting dynamics is often conjectured. Our experimental results showed that large random recurrent networks exhibit interesting functioning modes. Depending on a coupling parameter between neurons (e.g. the variance of the distribution of weights) the network is able to follow a wide spectrum of spontaneous behavior -- from the neural death (the initial stimulation is not enough to produce any further spiking activity) to the limit of maximal locking (some neurons fire all the time and the others never). In the intermediate states the average spiking activity grows. However we found out that the \emph{average} spiking activity reaches a steady-state. 
\par 
Thus, following basically the same ideas as \cite{amit97,fusi99}, we try to predict these behaviors when using large random networks. In this case, we make an independence hypothesis and use mean field techniques --  note that this so-called {\it mean field hypothesis} has been rigorously proved in a different neuronal network model \cite{moynot02}. More precisely, in our case, the connectivity weights will follow an independent identically distributed law and the neurons firing activities are supposed  independent. 
\par
After introducing the spiking neural model, we propose a mathematical formalism which allows us to determine (with some approximations) the probability law of the spiking activity. Since no other hypothesis than independence are used, a reinjection of the dynamics is needed. It leads expectedly to a massive use of recursive equations. However non intuitive, these equations are a solid ground upon which many conclusions can rigorously be drawn. 
\par
Fortunately, the solutions of these equations are as expected (and often taken for granted), that is the average spiking activity (and as a consequence the average frequency) reaches a steady state very quickly. Moreover, this steady state depends only on the parameters of the weight distribution. To keep the arguments simple, we detail the process for a weight matrix following a centered normal law. Extensions are proposed afterward for a non-zero mean and a sparse connectivity. All these results corroborate accurately with simulated neural networks data.  
 
\section{The neural model}%

The following  series of equations describe the discrete \emph{leaky integrate and fire} (I\&F) model we use throughout this paper \cite{tuckwell88}. Our network consists of $N$ all-to-all coupled neurons. Each time a given neuron fires, a synaptic pulse is transmitted to all the other neurons. This firing occurs whenever the neuron potential $V$ crosses a threshold $\theta$ from below. Just after the firing occurred, the potential of the neuron is reset to 0. Between a reset and a spike, the dynamics of the potential is given by the following (discrete) temporal equation~: 
\begin{equation}
  \label{equ:gene_disc}
  V_i(t) = \gamma V_i(t-1) +  I_i(t)   + \sum_{j=1}^{N} W_{ij} \delta (t - T_{j}-d_{ij}) 
\end{equation}
\par
The first part of the right hand side of the equation describes the leak currents ($0 \leq \gamma \leq 1$ $\gamma$ is the leak). Obviously, a value of 0 for $\gamma$ indicates that the neuron has no short term memory. On the other hand, $\gamma = 1$ describes a linear integrator.  
\par
The $W_{ij}$ are the synaptic influences (weights) and $I_i$ is an external input. $\delta(t)=1$ whenever $t=0$ and $0$ otherwise (Kronecker symbol). The $T_{j}$ are the times of firing  of a neuron $j$. $d_{ij}$ is an axonal delay and (as well as $T_{j}$) is a multiple of the sample discretization time.  
 The times of firing are defined formally for all neurons $i$ as~: 
\begin{equation}
\label{equ:lim_inf}
\lim_{\Delta t \rightarrow 0^{-}} V_i(T_i - \Delta t) = \theta 
\end{equation}
and so we define $T_i^n$ (the date of the n-th firing date) recursively as follow :
\begin{equation}
\label{equ:decharge}
 T_{i}^n = \inf (t ~ | ~ t> T_{i}^{n-1}, V_i(t)\geq \theta_i)
\end{equation}
We set that $T_{i}^{0} = -\infty $.
Moreover, once it has fired, the neuron's potential is reset to zero. 
\begin{equation}
\label{equ:reset}
  \forall n,\forall i,\quad \lim_{\Delta t \rightarrow 0^{+}} V_i({T_{i}^{n}} + \Delta t)=0
\end{equation}
\par
In this discrete description of an I\&F neuron, equations (\ref{equ:lim_inf}) and (\ref{equ:reset}) can be disturbing. They only mean that when computing $V_i(T_i^n + 1)$ we set 
$V_i(T_i^n) = 0$ in equation (\ref{equ:gene_disc}).

\par
In our study, we examine only spontaneous activity. It means that except at $t=0$ there is no external input : $\forall ~ i, \forall ~ t > 0, ~ I_i(t) = 0$. Moreover, given the above description, it is possible to allow great heterogeneity for various parameters. In order to simplify, we restrict ourselves in this paper to a synaptic weight heterogeneity leaving the other parameters constant. So $\forall ~ i, \forall ~j, ~d_{ij}=0$ (the axonal delay is 0). Similarly, $\forall ~i, ~\theta_i = \theta$ (same threshold for all neurons). Finally, suppose that the weights follow a centered normal law $N(0,\sigma^2)$ and let $\phi=\sigma\sqrt{N}$. 

\section{General Study}%

In this section we give a very general formulation of the distribution of the spiking activity defined as $X_t$ the numbers of firing neurons at a time step $t$ for an event.
We partition $X_t$ according to the instantaneous period of the neurons. That is we write : 
$$ 
X_t = X_{t}^{(1)} + \ldots + X_{t}^{(t-1)}
$$ 
when $X_t^{(k)}$ is the number of neurons who have fired at $t$ and $t-k$ but not in between.  Suppose that all neurons had a potential 0 at the starting process and only $X_0$ neurons were excited in order to make them fire, so $V_i(1)=\sum_{j=1}^{X_0}W_{ij}$. Thus using equation (\ref{equ:decharge}) we have: 
$$
X_1 = \sum_{i=1}^{N} \chi_{\{V_i(1)>\theta\}}
$$  
where $\chi_{\{V_i(1)>\theta\}}=1$ whenever $V_i(1)>\theta$ and 0 otherwise. Keeping on we have~: 
$$
X_2 = \sum_{i=1}^{X_1}\chi_{\{V_i(2)>\theta\}} + \sum_{i=1}^{N} \chi_{\{V_i(1)<\theta,V_i(2)>\theta\}}
$$ 
Indeed, the number of firing neurons at time step 2 are those that have fired twice (at $t=1$ and $t=2$ that is $X_2^{(1)} =\sum_{i=1}^{X_1}\chi_{\{V_i(2)>\theta\}}$ since the reset potential is 0). We need to add those that have not fired at $t=1$ ($X_2^{(2)} =\sum_{i=1}^N\chi_{\{V_i(1)<\theta,V_i(2)>\theta\}}$).

\par
Thus for $t$, taking into account the initial step, we have recursively~:
\begin{equation}
\label{equ:general_charge}
X_t = \sum_{u=1}^{t}\left( \sum_{i=1}^{\hat{X}_{t-u}} \chi_{\{V_i(t-u)<\theta,\ldots,V_i(t-1)<\theta,V_i(t)>\theta\}}\right)
\end{equation}
where $\hat{X}_{k}=X_k$ for $k\neq 0$ and $\hat{X}_0=N$. 

\par
Now assume that neurons dynamics are independent, we can calculate the expectation of $X_t$ (by the so-called {\it first Wald's identity})~:
\begin{equation}
\label{equ:esp_general_charge}
\E(X_t) = \sum_{k=0}^{t-1}\mathbb{E}(\hat{X}_{k})P(k,t)
\end{equation}
setting $P(k,t)=\mathbb{E}\left( \chi_{\{V_i(k+1)<\theta,\ldots,V_i(t-1)<\theta,V_i(t)>\theta\}} | V_i(k)=0 \right)$. 

The $P(k,t)$ are the expectation of a Bernoulli distribution. It leads that :
$$
\operatorname{Var}(\chi_{\{V_i(k+1)<\theta,\ldots,V_i(t-1)<\theta,V_i(t)>\theta\}}) = P(k,t)(1-P(k,t)).
$$ 
We are now able to retrieve the variance of $X_t$ (\emph{second Wald's identity}) : 

\begin{equation}
\label{equ:var_general_charge}
\operatorname{Var}(X_t) = \sum_{k=0}^{t-1}\left( \mathbb{E}(\hat{X}_k)P(k,t)(1-P(k,t))+\operatorname{Var}(\hat{X}_{k}){P(k,t)}^2 \right)
\end{equation}
More generally the moment generating function can be recursively computed :
\begin{equation}
\label{equ:car_general_charge}
\mathcal{G}_{X_t}(s) = \prod_{k=0}^{t-1} \mathcal{G}_{\hat{X}_{k}}(P(k,t)s + 1 - P(k,t))
\end{equation}

\section{Average Number Calculation}%

\par 
The equations (\ref{equ:esp_general_charge}) to (\ref{equ:car_general_charge}) are useful when we can estimate the $P(k,t)$ coefficients. We recall that $P(k,t)= \mathbb{E}\left( \chi_{\{V_i(k+1)<\theta,\ldots,V_i(t-1)<\theta,V_i(t)>\theta\}} | V_i(k)=0 \right)$. 
Thus the potential is a random sum of independent identically distributed normal laws (the weights). 
Unfortunately, this random sum is not, in general, a normal law itself.
Nevertheless, as it is proved in appendix \ref{app:sum}, (see equation \ref{lim:esperance}), when the number $N$ of neurons is large enough, and for a general class of distributions of the random variable $X_t$, we can write, for a neuron that has its potential $V_i(t) = 0$~:
\begin{equation}
\label{equ:gore}
V_i(t+1) \thicksim N\left(0,\mathbb{E}(X_t)\sigma^2\right) \Longrightarrow
P(V_i(t+1)>\theta) = \frac{1}{\sqrt{2\pi}}\int _{\frac{\theta}{\sqrt{\mathbb{E}(X_{t})}\sigma }}^{\infty} e^{-\frac{x^2}{2}}dx
\end{equation}
\par 
From now on we suppose that $\gamma \neq 0$. We need to know which are the neurons that have their potential $V_i(t) = 0$. When all neurons receive a charge from a non-zero number of neurons, the probability to have the potential equal to zero vanishes. Thus only the neurons that fired at time $t$ have this property (except for $t=0$ when it is true for \emph{all} the neurons). 
But taking into account a non-zero leak, we need to make here another independence assumption concerning the previous charges. Indeed the charge received by a neuron between $t$ and $t+k$ comes from  $\gamma^kX_t + \ldots + \gamma X_{t+k-1} + X_{t+k}$ ``neurons''. But in order to proceed further we assume that theses charges are independent. Note that when $\gamma$ is equal to zero this extra hypothesis is not needed. 
It leads to :
\begin{equation}
\label{eq:recurs_p}
P(k,t)  =  \prod_{m=1}^{t-k-1} \left[1-P\left(\sum_{j=1}^{l(k,m)}W_{ij}>\theta\right)\right]
P\left(\sum_{j=1}^{l(k,t-k)}W_{ij}>\theta\right)
\end{equation}
where we set
$$
l(k,m)=\sum_{i=k}^{m+k-1}\gamma^{m-k-i-1}\mathbb{E}(X_{i}).
$$
In order to lighten notations, we put $x_t = \frac{\mathbb{E}(X_t)}{N}$, and :
$$
p_\phi (y) = \frac{1}{\sqrt{2\pi}}\int _{\frac{\theta}{\sqrt{y}\phi }}^{\infty} e^{-\frac{x^2}{2}}dx
$$
We recall that we set $\phi=\sigma\sqrt{N}$. So equation (\ref{eq:recurs_p}) becomes :
\begin{equation}
P(k,t) = p_\phi\left(\sum_{i=k}^{t-1}\gamma^{t-i-1}x_{i}\right) 
\prod_{m=k+1}^{t-1}\left[1-p_\phi\left(\sum_{i=k}^{m-1}\gamma^{m-i-1}x_{i}\right)\right]
\end{equation}
Using the same notation we get the recursive computations of $\mathbb{E}(X_t)$:
\begin{equation} 
  \label{equ:nb_total}
  x_{t+1} = \sum_{k = 0}^t \hat{x}_k p_\phi\left(\sum_{i=k}^t\gamma^{t-i}x_{i}\right) 
\prod_{m=k+1}^t\left[1-p_\phi\left(\sum_{i=k}^{m-1}\gamma^{m-i-1}x_{i}\right)\right]
\end{equation}
where $\forall k>0, \hat{x}_k = x_k$ and $\hat{x}_0 = 1$.
\par
Moreover, we can compute the $P(k,t)$ recursively. Indeed, let $u_t^k = \sum_{i=k}^{t}\gamma^{t-i}x_{i}$ then $u_{t+1}^k = \gamma u_{t}^k + x_t$. Moreover :
\begin{equation}
P(k,t+1) = p_\phi(u_{t}^k ) \prod_{m=k+1}^{t}(1-p_\phi(u_{m-1}^k)) = p_\phi (u_{t}^k ) (\frac{1}{p_\phi (u_{t-1}^k)} -1 ) P(k,t)
\label{equ:pk_recur}
\end{equation}
with $P(t,t+1) = p_\phi(x_t)$ and $u_t^t = x_t$. $P(k,t)=0$ whenever $t\leq k$.

Equations (\ref{equ:nb_total}) and (\ref{equ:pk_recur}) are our main result. We can see that there is no more reference of the number of neurons. However, the algorithm grows exponentially with the time. 
\par 
Recalling a more classical spiking neurons formalism, we remark that equation (\ref{equ:esp_general_charge}) can be viewed as an integral over past charges of the form : 
\begin{equation}
x(t) = \int_{-\infty}^{t} x(\hat{t})P(\hat{t},t) d\hat{t}
\end{equation}
This is exactly Gerstner's formula \cite{gerstner01b} to compute spiking activity defined (using our notations) as~:
$$
x_t = \lim_{N\rightarrow \infty} \frac{1}{N} \sum_{n\geq 0} \sum_{i=0}^{N} \delta(t-T_i^{n})
$$  

\par
In the case $\gamma = 0$ we can deduce the result independently from the above equations. Indeed, since it means that at each time step the potential is reset to zero, we can directly write the probability of spiking of one neuron according to the number of neurons that have previously fired. Thus, with the same notations and hypothesis :
\begin{equation}
  \label{equ:nb_total_resistance}
  x_{t} = p_\phi(x_{t-1})
\end{equation}
But it is a special case of equation (\ref{equ:nb_total}) when $\gamma \rightarrow 0$. See appendix \ref{app:simple} for a detailed proof.

\section{Analysis}%
The equation (\ref{equ:nb_total}) is difficult to estimate. However some important things can be claimed. Following from the definition of the $P(k,t)$ we have for large enough $t$~:
\begin{equation}
\sum_{k=0}^{t} P(k,t) = 1
\end{equation}
It leads that $x_t$ is bounded and for large enough $t$ monotonic. Therefore, $x_t$ converges towards $x^{\star}$ when $t\rightarrow\infty$. $x^{\star} = 0$ (neural death) is an obvious solution. For high enough $\phi$ another fixed point exists bounded by $1/2$. Moreover, in this case, the $P(k,t)$ are close to a geometric distribution of parameter $p_{\phi}(x^{\star})$.  Due to the definition of $P(k,t)$, it leads to a geometric distribution of Inter-Spike Intervals that is~:
\begin{equation}
P(ISI = u) = P(t-u,t)
\end{equation}
It enables us to define a network frequency $f^{\star}$ defined as~:
\begin{equation}
f^{\star} = {(\E(ISI))}^{-1} = {(\E(u))}^{-1} = P_{\phi}(x^{\star})
\end{equation}
But if we define the network average frequency $\bar{F}(t)$ of a network over a period $T$ and at a given time $t$ by~:
\begin{equation}
\bar{F}(t) = \frac{1}{N} \sum_{i=1}^{N} \frac{1}{T} \sum_{k=0}^{T} \delta_i(t-k)
\end{equation}
where $\delta_i(t) = 1$ if the neuron $i$ has fired at time $t$ and $\delta_i(t)=0$ otherwise (in other words $\delta_i(t) = \sum_{n\geq0}\delta(t-T_i^n)$). Switching the sum symbol gives~:
\begin{equation}
 \bar{F}(t) = \frac{1}{T} \sum_{k=0}^{T} \frac{1}{N} \sum_{i=1}^{N} \delta_i(t-k) = \frac{1}{T} \sum_{k=0}^{T} \frac{1}{N} X_{t-k}
\end{equation}
for this realization of the distribution. Taking the expectation leads to~:
\begin{equation}
\bar{f}(t) = \E(\bar{F}(t)) = \frac{1}{T} \sum_{k=0}^{T} x_{t-k}
\end{equation}
It means that the average frequency (on a time window $T$) is the average of the spiking activity (over a period $T$). Thus, when $t\rightarrow \infty$ it leads to~:
\begin{equation}
\bar{f}(t) = x^{\star}
\end{equation}
Due to discrete timing, we  generally don't have $f^{\star}=\bar{f}$. Instead we have $f^{\star}<\bar{f}$. It gives~:
\begin{equation}
p_{\phi}(x^\star) \leq x^{\star}
\end{equation}
The inequality becomes an equality if $\gamma = 0$. This is the case we now study. 

\subsection{Simple case}
\par
If we consider the case $\gamma = 0$, we recall that $x_t = p_\phi(x_{t-1})$. So $x^{\star}=p_{\phi}(x^{\star})$. 
\par
 A solution $x^\star \neq 0$ exists when $p_\phi(x)$ crosses the line $y=x$ and is stable if and only if $p_\phi'(x^\star)<1$ (here $p_\phi'$ is positive on all the positive line). If it exists $x$ such that $p_\phi(x)>x$ then $p_\phi(x)=x$ has only two solutions. The first (lowest) one is an unstable fixed point and the other is a stable fixed point. So if $x_0$ is above the lowest fixed point, the average number of neurons converges toward $x^\star$. In the other case, it converges to 0 (i.e neural death). We can derive a sufficient condition for the convergence to zero (see appendix \ref{app:derive} for details)~:
\begin{equation}
 \phi < \left(\frac{2e}{3}\right)^{\frac{3}{4}}\pi^{\frac{1}{4}}\theta \approx 2.08 \theta
\end{equation}

\subsection{Previous charges independence}
In the case $\gamma \neq 0$, we now need the independence of charges hypothesis. However in the general case, we allowed the potential to have strong negative values. More than to be biologically wrong, it dramatically impedes the independence hypothesis. Indeed, some neurons with very low potential will never fire whatever happens. 

In order to take this into account, we make a (biologically plausible) assumption : the potential is not allowed to decrease under a minimal value $v_{min}$. This leads us to reconsider the charge function :
$$
P(k,t)=\mathbb{E}(\chi_{\{V_i(t-k+1)<\theta,\ldots,V_i(t-1)<\theta,V_i(t)>\theta\}})
$$

We recall that under the hypothesis of independence and since we have a normal law for the weights it leads to :
\begin{equation}
P(k,t) = \hat{p}_\phi\left(\sum_{i=k}^{t-1}\gamma^{t-i-1}x_{i}\right) 
\prod_{m=k+1}^{t-1}\left[1-\hat{p}_\phi\left(\sum_{i=k}^{m-1}\gamma^{m-i-1}x_{i}\right)\right]
\end{equation}
where $\hat{p}_\phi$ is the new probability function we need to compute. 

Let's assume that a neuron $i$ has taken a charge $\mathcal{C}$ in the previous times and is subject to a charge $x_t$ at time $t$. The probability that the total charge exceeds the threshold must be split in two cases : the potential occurred by $\mathcal{C}$ is below $v_{min}$ or not. In the first case (below $v_{min}$), the charge will be $x_t$ on a potential $v_{min}$ with probability $p=P(\Phi<v_{min})$, where $\Phi \thicksim N(0,\mathcal{C}\sigma^2)$. In the other case it will be $\mathcal{C}+x_t$ (with probability $1-p$). So the resulting probability will be :
\begin{eqnarray*}
\label{equ:with_prob}
P(V_i(t)>\theta) &=& P(\Phi>\theta) \mbox{~where~} \Phi \thicksim ( pN(v_{min},x_t\sigma^2) + (1-p)N(0,(\mathcal{C}+x_t)\sigma^2))\\
& =& P(\Phi>\theta)\mbox{~where~} \Phi \thicksim N(pv_{min},((1-p)\mathcal{C}+x_t)\sigma^2), 
\end{eqnarray*}
We can find the $P(k,t)$ recursively noting that the probability $p$ depends on previous $x_t$. 
To simplify, we suppose that $v_{min}=0$ (i.e the neuron cannot have negative values) it is even simpler. In this case, whatever the charge is, the probability $p$ is always $1/2$. So equation (\ref{equ:with_prob}) becomes :
\begin{equation}
P(V_i(t)>\theta) = P(\Phi \thicksim N(0,(\frac{\mathcal{C}}{2}+x_t)\sigma^2), \Phi>\theta)
\end{equation}
It acts as if the decay rate was divided by two. Thus into equation (\ref{equ:nb_total}) it leads to :
\begin{equation} 
  \label{equ:nb_total_half}
  x_{t+1} = \sum_{m = 0}^{t} \hat{x}_mp_\phi(\sum_{i=0}^{t-m}{(\frac{\gamma}{2})}^i x_{t-i})\prod_{j=0}^{t-m-1}(1-p_\phi(\sum_{k=0}^{j}{(\frac{\gamma}{2}})^k x_{j-k+m})) 
\end{equation}
where $\forall m>0, \hat{x}_m = x_m$ and $\hat{x}_0 = 1$.

\section{Results and comparison}
We conducted extensive numerical simulations to confront our formulas. The results we present here were made with 500 random networks of 1000 I\&F neurons. The starting stimulated neurons were chosen randomly for each network tested -- each neuron was scanned and set to fire with a probability $x_0$. $x_0$ was chosen between $0.05$, $0.30$, $0.5$, $0.8$ and $1.0$. To be in concordance with equation (\ref{equ:nb_total_half}) we set $v_{min} = 0$. 

\subsection{Asymptotic Spiking Activity Comparison}
\begin{figure}[htb]
\begin{center}
\includegraphics[width=8cm]{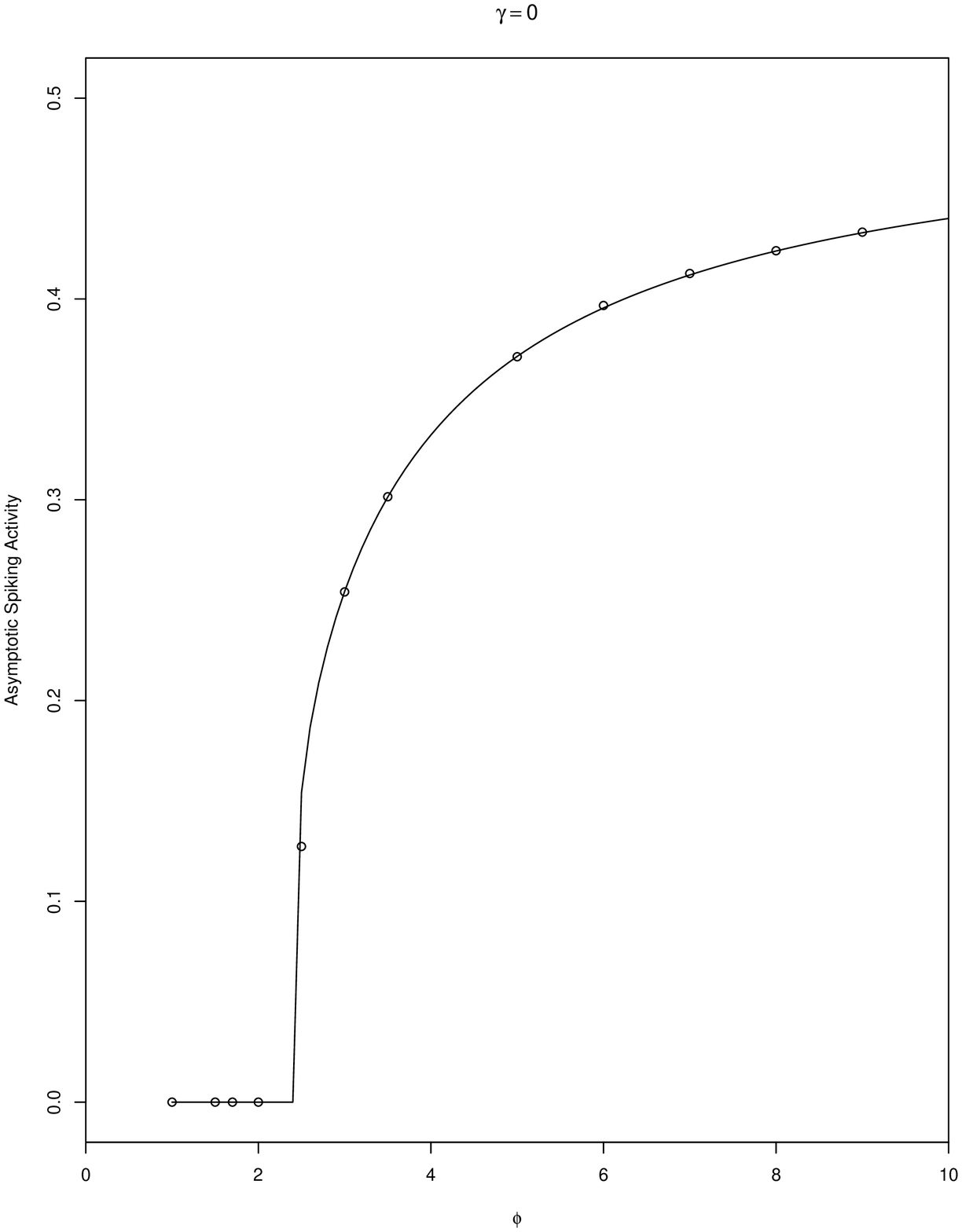}
\includegraphics[width=8cm]{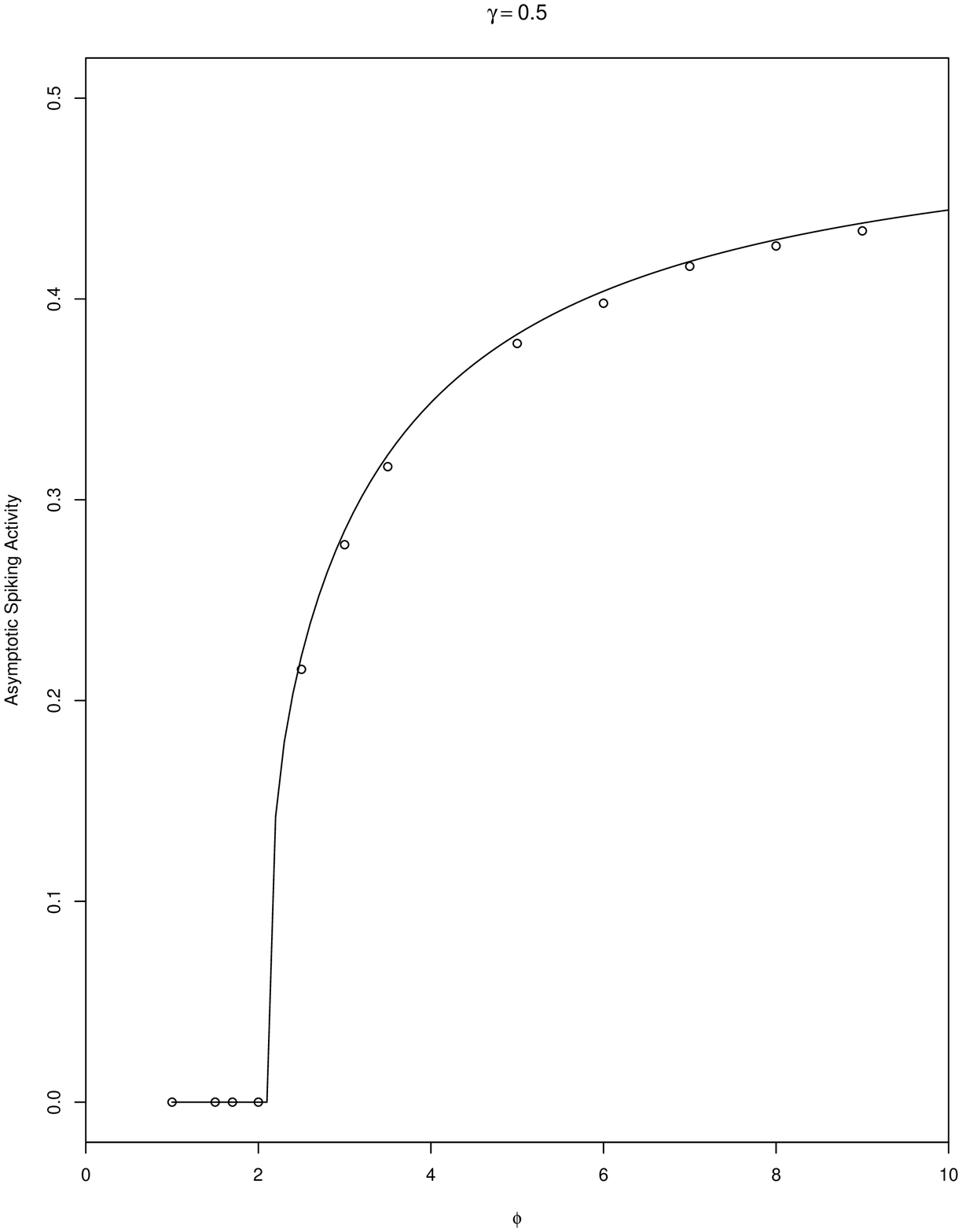}
\includegraphics[width=8cm]{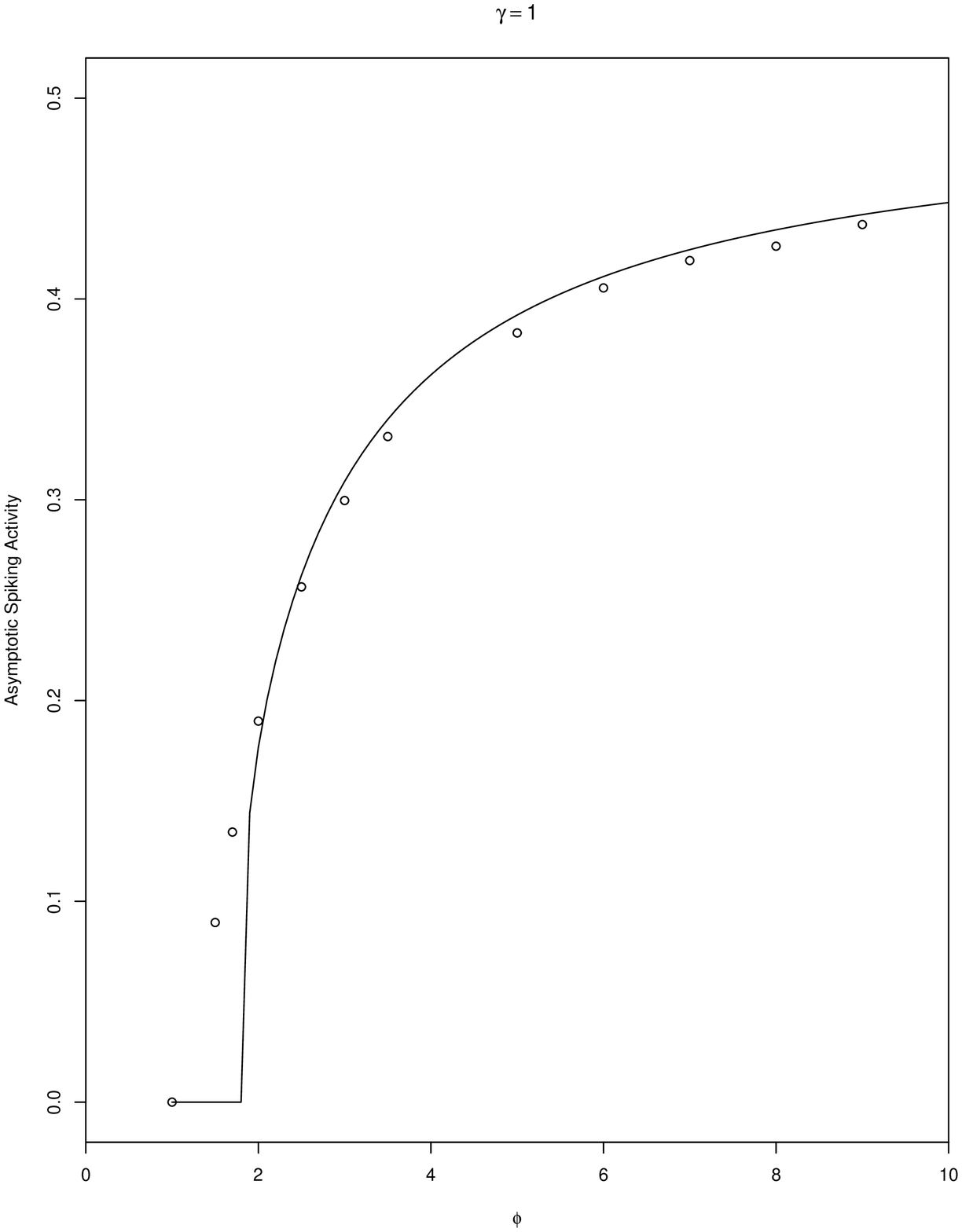}
\caption{Asymptotic limit as a function of $\phi$ for $\gamma \in \{0.0,0.5,1.0\}$. Theoretical results are displayed with a plain line, while experimental data points are circles. Parameters:~ $N ~=~ 1000$,$~\theta~=~1$}
\label{fig:asym_mode}
\end{center}
\end{figure}

We recorded the limit of $x_t$ according to different values of $\phi$ and for  $\gamma = 0.0,0.5,1.0$. The results of the theoretical and experimental simulations are displayed in figure \ref{fig:asym_mode}. The limit of $x_t$ is an increasing function of $\phi$. Moreover $\lim_{\phi\rightarrow +\infty} \left( \lim_{t \rightarrow +\infty} x_t\right) = 1/2$. As easily seen, the higher $\phi$ and lower $\gamma$ give an extremely good prediction compared to experimental data. A high $\gamma$ always gives a slight overestimation in the case of high $\phi$. Indeed, the independence of charge never holds. On the other hand, when this hypothesis is rigorously true (i.e $\gamma = 0$) the prediction is strikingly accurate. 
For small values of $\phi$, the prediction was correct also. However it exists an interval of intermediate values of $\phi$ (that is here between $1.5$ and $2.0$) where the prediction completely fails. Besides we shall see in the next section that this is also the case for the transient period. Indeed, for small enough coupling, the independence of the dynamics of neurons is not true. We expect that the length of this interval tends towards zero when $N$ grows infinitely. 
\subsection{Transient period}

\begin{figure}[htb]
\begin{center}
\includegraphics[width=18cm]{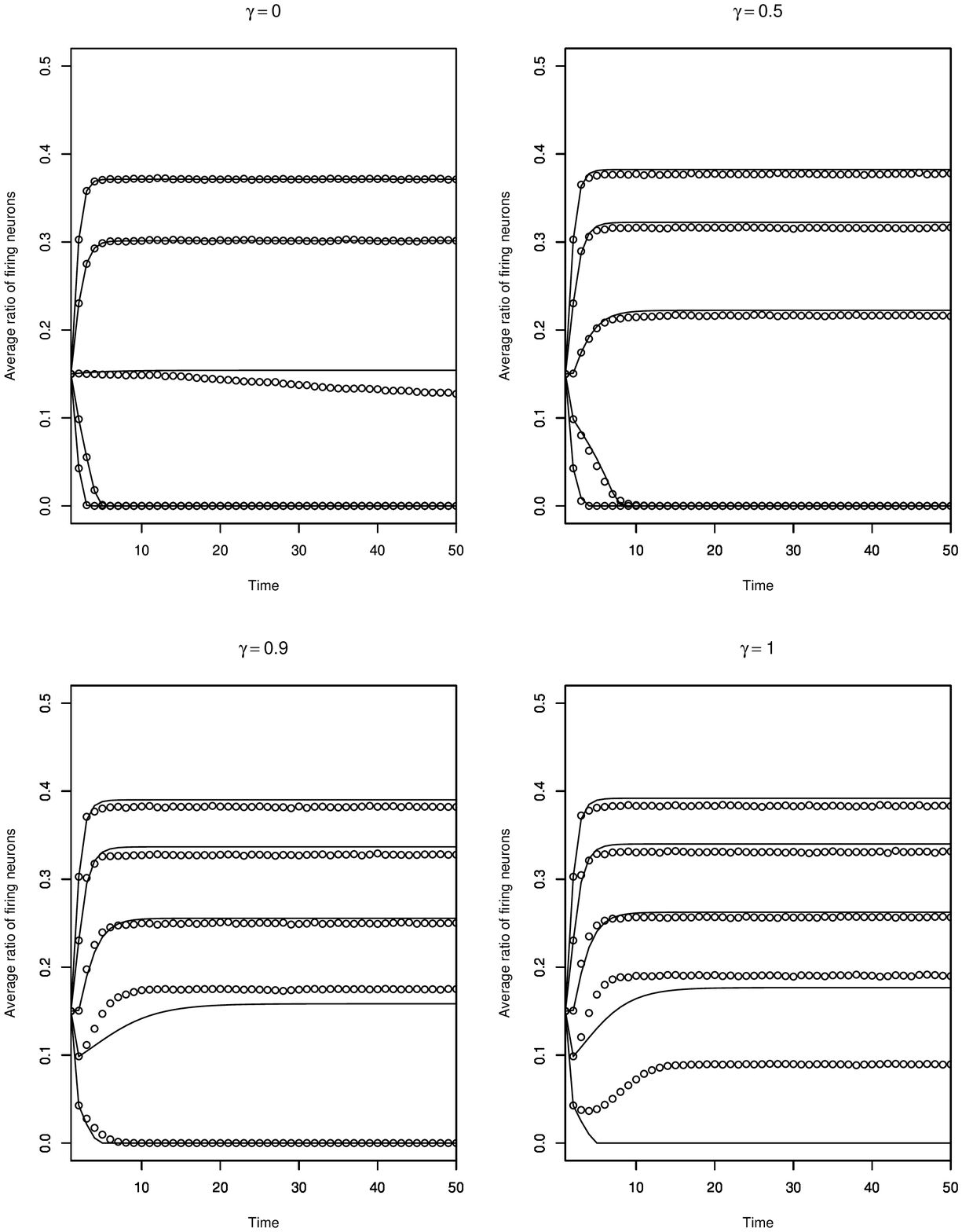}
\caption{Results for various values of $\phi \in \{1.5,2.5,3.5,5.0\}$ and $\gamma \in \{0.0, 0.5,0.9,1.0\}$. We used the same $x_0 = 0.15$ for each simulations. Theoretical results are displayed with a plain line, while experimental data points are circles. Parameters:~ $N ~=~ 1000$,$~\theta~=~1$}
\label{fig:aper_mode}
\end{center}
\end{figure}

We recorded the evolution of the spiking activity for the transient period before reaching the asymptotic limit for various values of $\phi$ and $\gamma$ during the first 50 time steps. The results are displayed on figure \ref{fig:aper_mode}. As in the previous section, a higher $\gamma$ gives a slight overestimation but is completely correct when $\gamma$ draws near zero. Moreover, we retrieve the same interval of $\phi$ where the neurons dynamics independence fails (and as a consequence the transient). These results show that the steady-state is reached very quickly (less than 10 time steps).   

\section{Generalization}
We supposed to make the arguments clear that the weight distribution followed a very simple law -- a centered normal law. However, we do not have to restrict ourselves to this limit. Indeed, we look at two possible extensions of spiking activity equations. We show that it is possible to use a non-zero mean for the distribution as well as a sparse weight matrix in a certain sense. The results obtained showed the same characteristics.

\subsection{Non-zero mean}
\begin{figure}[htb]
\begin{center}
\includegraphics[width=18cm]{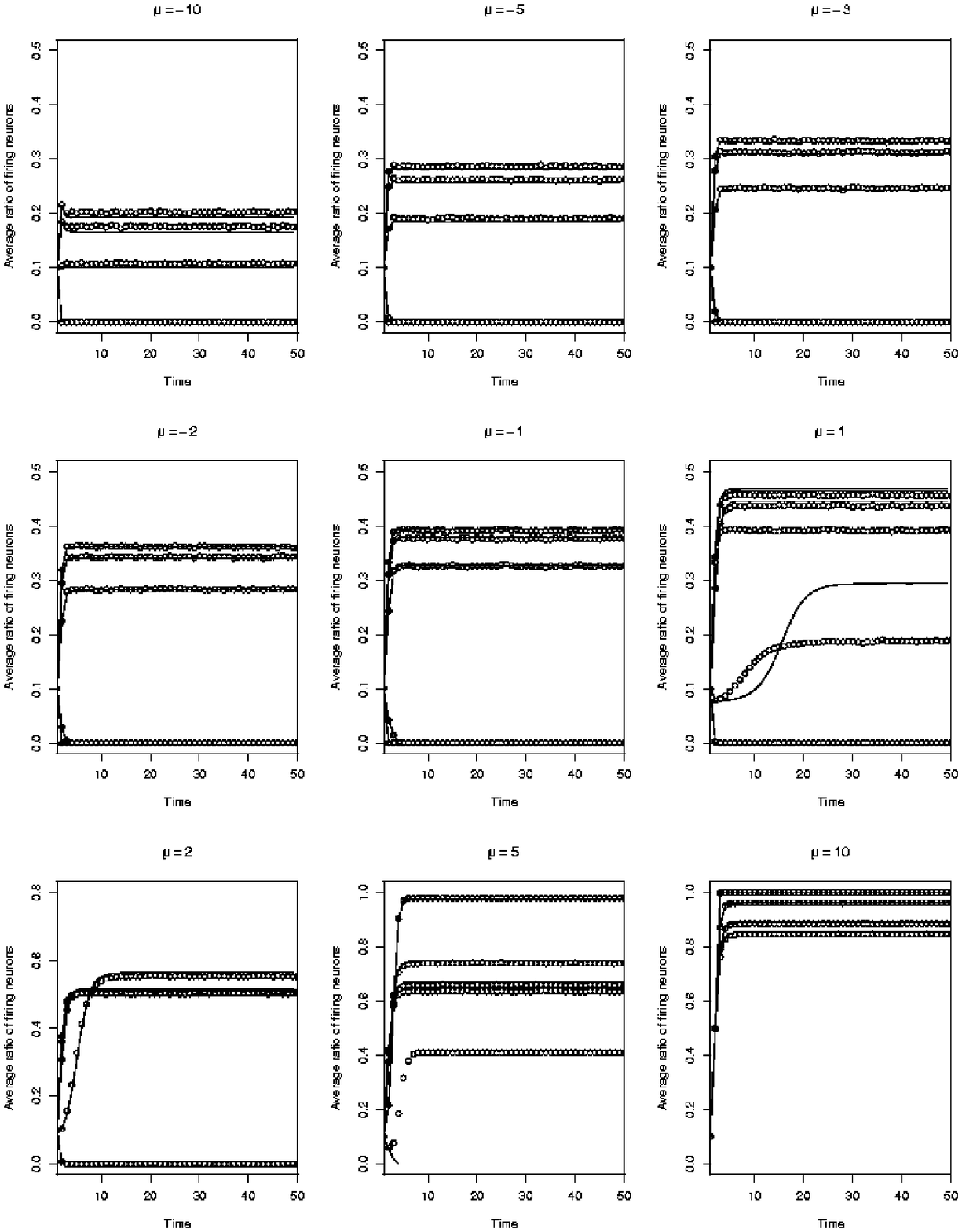}
\caption{Transient with different mean $\mu$ for  $\phi \in \{1.0,2.0,5.0,7.0,8.0\}$. Theoretical results are displayed with a plain line, while experimental data points are circles. Parameters:~ $N ~=~ 1000$,$~\theta~=~1$,$~\gamma~=~0.5$,$~x_0~=~0.1$.}
\label{fig:mu}
\end{center}
\end{figure}

In fact taking a non-zero mean is very easy. Indeed, provided that all the hypotheses remain valid, we can insert the mean into the definition of $p_\phi$ which becomes $p_{\phi,\mu}$ defined as~:
$$
p_{\phi,\mu} (y) = \frac{1}{\sqrt{2\pi}}\int _{\frac{\theta-\mu y}{\sqrt{y}\phi }}^{\infty} e^{-\frac{x^2}{2}}dx
$$
which corresponds to a normal weight distribution $N(\frac{\mu}{N},\frac{\phi}{\sqrt{N}})$. 
In order to relate this to a very simple case, we can look at what happens when $\phi\rightarrow 0$ -- that is an homogeneous network. It leads to $p_{\phi,\mu}(y) = 0$ when $\mu y<\theta$, $p_{\phi,\mu}(y) = 1$ when $\mu y>\theta$ and finally $p_{\phi,\mu}(y) = 1/2$ when $\mu y=\theta$ . All depends on the value $\frac{\mu x_0}{\theta}$. \\
\textbf{Case $\frac{\mu x_0}{\theta}<1$} In this case, $x_1 = p_{\phi,\mu}(x_0) = 0$ and it converges to neural death that is $\forall ~t>0~ x_t~=~0$~.\\
\textbf{Case $\frac{\mu x_0}{\theta}>1$} Here, $x_1 = p_{\phi,\mu}(x_0) = 1$ leads to $\forall ~t>0~ x_t= 1$. All neurons fire at the same time and with the highest frequency (one spike per time step).\\
\textbf{Case $\frac{\mu x_0}{\theta}=1$} Now, $x_1 = p_{\phi,\mu}(x_0) = 1/2$. Suppose $x_0<1/2$ then $p_{\phi,\mu}(x_1)=1$ leading $x_2 = x_1 p_{\phi,\mu}(x_1) +  p_{\phi,\mu}(\gamma x_0 + x_1)(1- p_{\phi,\mu}(x_0)) = 1/2\times1 + 1\times1/2=1$ leading to the case $\frac{\mu x_0}{\theta}>1$. On the other hand if $x_0>1/2$ then $p_{phi,\mu}(x_1)=0$ leading to case $\frac{\mu x_0}{\theta}<1$. It left us with $x_0=x_1=1/2$. In this case, $x_2 = 1/2$ as for all other $x_t$. This result can be strange if we forget that $\phi\ll1$ but $\phi\neq=0$ (if not the weight independence is no longer true) leading an average value between the two first cases (hence $x_t = 1/2$). \\
The general behavior does not change with non-zero mean. The convergence is very fast and the transient are well predicted as shown in figure \ref{fig:mu}. The only difference with $\mu>0$ is that $x^{\star}$ can be well above $1/2$ and ultimately reach 1. 

\subsection{Sparse connectivity}
\begin{figure}[htb]
\begin{center}
\includegraphics[width=18cm]{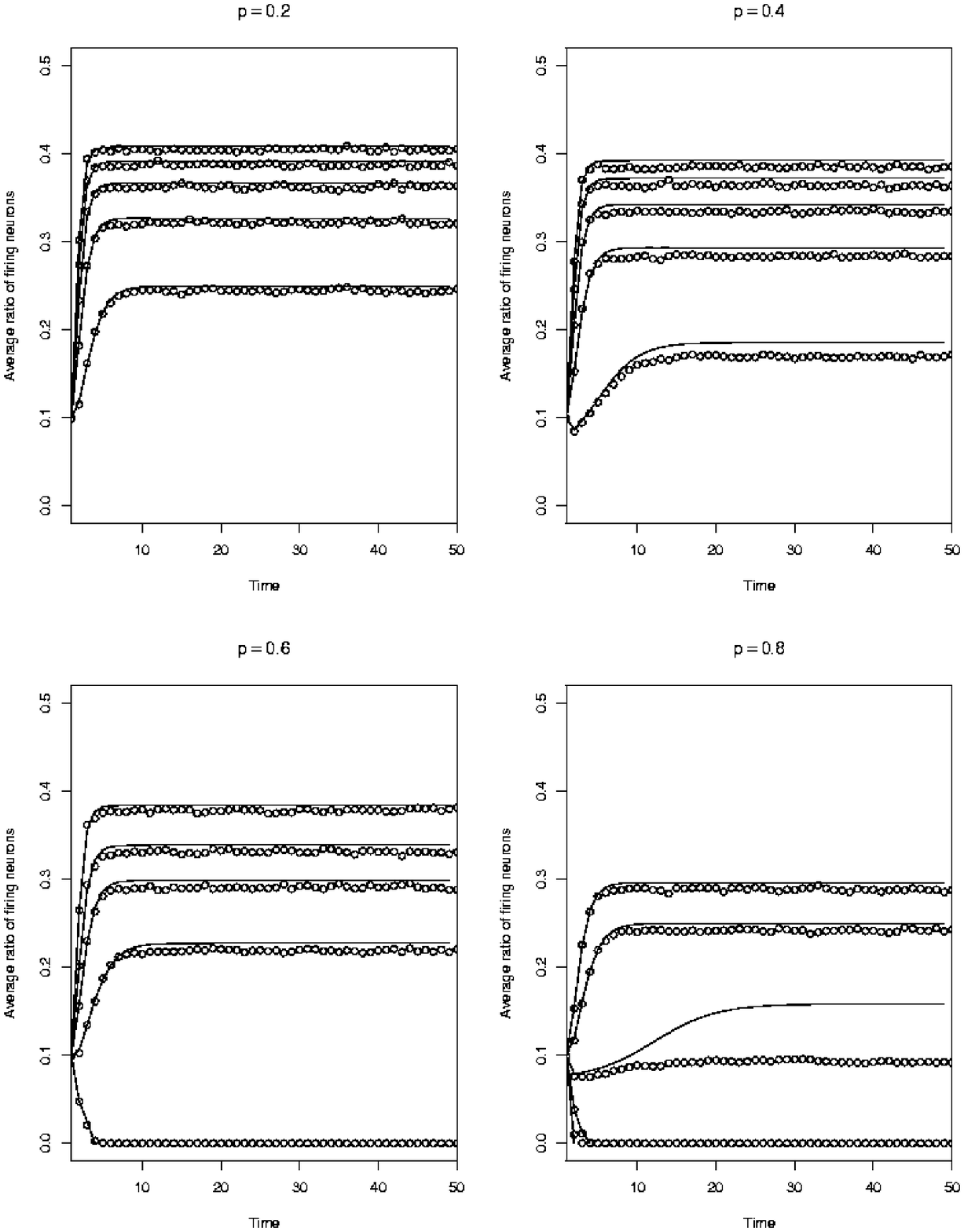}
\caption{Transient with different sparse parameter $p$ (we recall that $p=0$ means a zero matrix while $p=1$ is a pure normal distribution) for  $\phi \in \{3.0,4.0,5.0,6,0,7.0\}$. Theoretical results are displayed with a plain line, while experimental data points are circles. Parameters:~ $N ~=~ 1000$,$~\theta~=~1$,$~\gamma~=~0.5$,$~x_0~=~0.1$.}
\label{fig:sparse}
\end{center}
\end{figure}

A pure normal distribution is unlikely to occur in biological neurons. Most of the time, neurons are linked to a certain amount of neurons in the network. It can be viewed using a sparse weight matrix (a matrix with zero coefficient). In order to take this into account we compute sparse matrix as follows. A weight $w$ has a probability $p$ to be zero and a probability $1-p$ to follow a normal law $N(\mu/N,\phi/\sqrt{N})$. As in the previous section it leads to a new $\hat{p}_{\phi,\mu}$ function. When calculating the charge, it came from ``$X$'' neurons leading to a sum of $X$ normal laws. In the case of sparse matrix, it reduce to $(1-p)X$ neurons. So our new function $\hat{p}_{\phi,\mu}$ becomes:
$$
\hat{p}_{\phi,\mu}(y) = {p}_{\phi,\mu}((1-p)y)
$$
It remains to insert this new function into equation (\ref{equ:nb_total_half}). As shown in figure \ref{fig:sparse}, the general behavior is not changed when introducing sparsity.

\section{Discussion}%
Independence hypothesis can be a very powerful way to approach random networks of spiking neurons. So far, we proved that the ``coupling factor'' ($\phi = \sigma\sqrt{N}$) can characterize the average spiking activity. For instance whatever the initial stimulation (provided it is strong enough and independent of the neurons), the networks average spiking activity reaches the same steady-state. Since the death locking is also a possible steady-state, in dynamical systems terminology, depending on the value of $\phi$,  we exhibited a bifurcation. 
\par
In addition, this steady-state only depends on the coupling parameter (for a fixed type of neurons i.e. same leak). This dependence (if not obvious) is increasing and converges to $1/2$ when $\phi$ grows larger. That is a limit state where half the neurons that are active fire all the time and the remaining stay silent. That is an \emph{extreme locking}. Indeed, all neurons are periodic (either 1 or $\infty$) and those that fire do it synchronously (that is all the time). 
\par 
The transient phase was also well determined -- very quick convergence to the steady-state. We were also able to derive mathematically a sufficient condition for death locking (in a very simple case though). 
\par
However, as showed for a certain range of coupling factor, independence is deemed to fail and consequently the prediction -- both in transient and asymptotically. Indeed, intuitively when a coupling value is intermediate, neurons can no longer behave as if it did not matter who fired. It seems somehow that this range of ``failed independence'' grows thin with increasing $N$ (at the thermodynamical limit).
\par
This shows that a spontaneous regime can be self-sufficient. It means that networks can afford discontinuous inputs without losing their internal activity. In other words, it is theoretically possible to observe dynamic cerebral areas without any permanent external stimulation. 

\newpage
\bibliographystyle{apacitex}
\bibliography{spike_dynamics}
\newpage

\begin{appendix}
\appendix
\section{Random sum of random variables}%
\label{app:sum}
Let us first prove a general result, which will have an application for the neuronal potential, written as a random sum of i.i.d. random variables.

Let $f$ be a function so that $\lim_{k \to \infty}f(k)=\alpha\in \mathbb{R}$, $(p_k^N)_{(k,N)\in \mathbb{N}^2}$ a sequence satisfying $\forall k\in \mathbb{N}$, $\lim_{N\to \infty}p_k^N=0$ and $\sum_{k=1}^Np_k^N=1$.
Let now define $g(N)=\sum_{k=1}^N p_k^N f(k)$, and prove that 
\begin{equation}
\label{lim:esperance}
\lim_{N\to \infty}g(N)=\alpha.
\end{equation}
$\forall \epsilon>0, ~ \exists N_0\in \mathbb{N}, ~ \forall k>N_0, ~ |f(k)-\alpha|<\frac{\epsilon}{2}$. So we can write
$$
|g(N)-\alpha|\leq \left|\sum_{k=1}^{N_0}p_k^N(f(k)-\alpha)\right|+\left|\sum_{k=N_0}^Np_k^N(f(k)-\alpha)\right|
\leq \left|\sum_{k=1}^{N_0}p_k^N(f(k)-\alpha)\right|+\frac{\epsilon}{2}.
$$
$N_0$ being fixed, it just remains, to complete the proof, to get a rank $N_1$ so that
$$
\forall N>N_1, \quad \left|\sum_{k=1}^{N_0}p_k^N(f(k)-\alpha)\right|<\frac{\epsilon}{2}.
$$

{\bf Application :} let $X^{(N)}$ be a sequence of random variables on $[1,N]$ so that
$$
\lim_{N\to \infty}\mathbb{E}(X^{(N)})=\infty \mbox{~~and~~} \forall k, ~ \lim_{N\to \infty}P(X^{(N)}=k)=0,
$$
which is satisfied for instance when $X^{(N)}\sim N(\frac{N}{2},\sigma^2)$ or when $X^{(N)}$ is uniform on $[1,N]$.
In this case, if we set $p_k^{(N)}=\mathbb{P}(X^{(N)}=k)$ we can write $g(N)=\mathbb{E}(f(X^{(N)}))$, and equation (\ref{lim:esperance}) yields
\begin{equation}
\lim_{N\to \infty}  \mathbb{E}(f(X^{(N)})-f(\mathbb{E}(X^{(N)})) =0.
\end{equation}
Equation \ref{equ:gore} derives from the case
$$
f(k)=\frac{1}{\sqrt{2\pi}}\int_{\frac{1}{\sqrt{k}\sigma}}^{+\infty}e^{-\frac{x^2}{2}}\, dx.
$$

\section{Simple Case}%

\label{app:simple}
We prove here that substituting  $\gamma=0$ in equation (\ref{equ:nb_total}) give the equation~:
\begin{equation*}
  x_{t} = p_\phi(x_{t-1})
\end{equation*}
It yields :
\begin{equation*} 
x_{t+1} = \sum_{m = 0}^{t} \hat{x}_mp_\phi(x_{t})\prod_{j=m}^{t-1}(1-p_\phi(x_{j}))
\end{equation*}
For $t=0$, it gives $x_1=p_\phi(x_0)$. Using the recurrence hypothesis $\forall m=1\ldots t$, $x_m = p_\phi(x_{m-1})$ we find that :
\begin{eqnarray*}
x_{t+1} &=& \sum_{m = 0}^{t} p_\phi(x_{m-1})p_\phi(x_{t})\prod_{j=m}^{t-1}(1-p_\phi( x_{j}))\\
&=&p_\phi(x_t)\sum_{m = 0}^{t}p_\phi(x_{m-1})\prod_{j=m}^{t-1}(1-p_\phi(x_{j}))\\
&=&p_\phi(x_t)v_{t-1}\\
\end{eqnarray*}
Is is now enough to note that 
\begin{eqnarray*}
  v_{t-1} &=& p_\phi(x_{t-1}) + (1-p_\phi(x_{t-1}))\sum_{m = 0}^{t-1}p_\phi(x_{m-1})\prod_{j=m}^{t-2}(1-p_\phi(x_{j})))\\
  &=& p_\phi(x_{t-1})+(1-p_\phi(x_{t-1}))v_{t-2}\\
\end{eqnarray*}
Since $v_0 = (1-p_\phi(x_{0})) +p_\phi(x_{0}) = 1$, we have $v_{t}=1$ for all $t$. We showed by recurrence that we get equation (\ref{equ:nb_total_resistance}).

\section{Sufficient condition for death}%

\label{app:derive}
Let's go back to 
$$ 
p_\phi (y) = \frac{1}{\sqrt{2\pi}}\int _{\frac{\theta}{\sqrt{y}\phi }}^{\infty} e^{-\frac{x^2}{2}}dx,
$$
for $y\in [0,1]$.
Taking the derivative over $y$ we have :
$$ 
p'_\phi(y) = \frac{\theta}{2\sqrt{2\pi}\phi y^{3/2}}e^{-\frac{\theta^2}{2y\phi^2}}.
$$
We see immediately that $p'_\phi(y)\geq 0$ so $p_\phi$ is increasing. So, stable non-zero fixed points should appear for a value of $y$ that crosses the line $y=x$ from above. Then a sufficient condition is that $\forall y $ ~: 
$$ 
p'_\phi(y) < 1.
$$ 
Let $z = \frac{1}{y}$, $\tau = \frac{\theta}{\sqrt{2}\phi}$ and $g(z)=p'_\phi(\frac{1}{z})$ then~:
$$
g(z) = \frac{\tau z^{3/2}}{2\sqrt{\pi}}e^{-\tau^2z}
$$
Taking the derivative of $g(z)$ yields~:
$$
g'(z)=
\frac{\tau z^{1/2}}{2\sqrt{\pi}}e^{-\tau^2z}\left(\frac{3}{2}-\tau^2z\right)
$$
So $g'(z)$ has the same sign as $\frac{3}{2\tau^2}-z$ and since $g'(0) = g'(+\infty) = 0$, it leads that the maximum of $g(z)$ is obtain for $z= \frac{3}{2\tau^2}$. But $z = \frac{1}{y}$ thus $z\in [ 1,+\infty [ $. 
It leads that $$\max_{y\in [0,1]}p'_\phi(y)=\max_{z\in [1,+\infty[}g(z) = \max\left(g\left(\frac{3}{2\tau^2}\right),g(1)\right)$$
So a sufficient condition for zero to be the only fixed point becomes :
$$g\left(\frac{3}{2\tau^2}\right) = \frac{\tau{(\frac{3}{2\tau^2})}^{3/2}}{2\sqrt{\pi}}e^{-\tau^2\frac{3}{2\tau^2}}
= \left(\frac{3}{2e}\right)^{\frac{3}{2}}\frac{1}{2\sqrt{\pi}\tau^2}<1$$ 
Since $\tau = \frac{\theta}{\sqrt{2}\phi}$ we get :
$$
 \phi < \left(\frac{2e}{3}\right)^{\frac{3}{4}}\pi^{\frac{1}{4}}\theta
$$

\end{appendix}

\end{document}